\documentclass[11pt,a4paper]{article}
\usepackage[hyperref]{emnlp-ijcnlp-2019}
\usepackage{times}
\usepackage{latexsym}

\usepackage{graphicx}
\usepackage{url}
\usepackage{mathtools}

\aclfinalcopy % Uncomment this line for the final submission

\title{Taxonomical hierarchy of canonicalized relations from multiple Knowledge Bases}

\author{Akshay Parekh \\
  IIT Guwahati \\
  Guwahati, Assam, \\
  India \\
  \And
  Ashish Anand \\
  IIT Guwahati \\
  Guwahati, Assam, \\
  India \\
  \And
  Amit Awekar \\
  IIT Guwahati \\
  Guwahati, Assam, \\
  India \\
  }

\date{}

\begin{document}
\maketitle

\begin{abstract}
    This work addresses two important questions pertinent to Relation Extraction (RE). First, what are all possible relations that could exist between any two given entity types? Second, how do we define an unambiguous taxonomical (\texttt{is-a}) hierarchy among the identified relations? To address the first question, we use three resources Wikipedia Infobox, Wikidata, and DBpedia. This study focuses on relations between \textit{person}, \textit{organization} and \textit{location} entity types. We exploit Wikidata and DBpedia in a data-driven manner, and Wikipedia Infobox templates manually to generate lists of relations. Further, to address the second question, we canonicalize, filter, and combine the identified relations from the three resources to construct a taxonomical hierarchy. This hierarchy contains 623 canonical relations with the highest contribution from Wikipedia Infobox followed by DBpedia and Wikidata. The generated relation list subsumes an average of $85\%$ of relations from RE datasets when entity types are restricted \footnote{Resources for the relation hierarchy is available at \url{https://github.com/akshayparakh25/relationhierarchy}}.
\end{abstract}

\section{Introduction}
\label{sec:intro}
\begin{figure}
        \centering
        \includegraphics[width=0.5\textwidth]{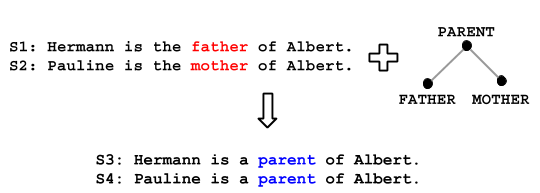}
        %\caption{Example showing \textit{father} and \textit{mother} shares \texttt{is-a} relation with \textit{parent}.}
        \caption{With the taxonomical hierarchy, and the relations present in the sentences S1 and S2, we can infer relations shown in sentences S3 and S4.}
        \label{fig:example}
    \end{figure}

    Relations mentioned in unstructured texts often share taxonomical (\texttt{is-a}) association with other relations. For example, in figure \ref{fig:example} relations \textit{father} and \textit{mother} shares taxonomical relation with \textit{parent}. By virtue of this relation, entities \texttt{Hermann} and \texttt{Pauline} also have \textit{parent} relation with the entity \texttt{Albert} which is also true in real-world. But such inference is hard to extract from the existing relation extraction (RE) resources as they fail to answer two following questions: \textit{first}, what are all possible relations that could exist between entities? \textit{Second}, how do we obtain an unambiguous taxonomical hierarchy between the identified relations?
    
    %Often these relations share taxonomical (\texttt{is-a}) association with other relations. Figure \ref{fig:example} shows an example of two relations \textit{father} and \textit{mother} sharing taxonomical relation with \textit{parent}. By virtue of this relation, entities \texttt{H Einstein} and \texttt{P Koch} also have \textit{parent} relation with the entity \texttt{A Einstein}. Such inference is hard to extract from the existing relation extraction (RE) resources.  
    %Also, information shared in those texts incorporates sophisticated information between entities and are often taxonomically similar although semantically different. For example, \textit{father} and \textit{mother} are semantically different but shares \texttt{is-a} relation with \textit{parent}.
    %This requires answers of the two questions: \textit{first}, what are all possible relations that could exist between entities? \textit{Second}, how do we obtain an unambiguous taxonomical hierarchy between the identified relations?
    
    %\textbf{alternate}
    
    Available RE resources show the following bottlenecks: \textit{limited relations}, \textit{absence of canonical relations}, and \textit{absence of hierarchy in them}. The first limitation is because of the pre-defined handcrafted or corpus-dependent relations list \cite{mitchell2005ace, hendrickx2009semeval}. To scale the number of relations a few datasets \cite{riedel2010modeling,hao2018fewrel} use a single KB to obtain a potential list of relations. As there is no standard nomenclature and mapping followed among KBs, restriction to a single KB leads to the second bottleneck. Even though KBs like Wikidata \cite{vrandevcic2014wikidata} and DBpedia \cite{lehmann2015dbpedia} incorporate deep hierarchical ontologies, that do not explicitly address relations and extracting relation hierarchy from them is challenging.
    \begin{figure*}[t]
        \centering
        \includegraphics[width=0.9\textwidth]{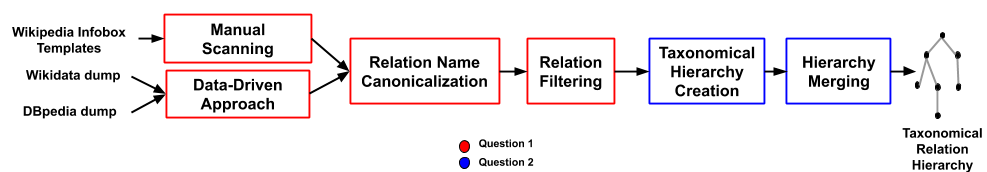}
        \caption{Overview of hierarchy creation steps. Question 1: What are all possible relations that could exist between any two given entity types? Question 2: How do we define an unambiguous taxonomical (\texttt{is-a}) hierarchy among the identified relations?}
        \label{fig:flowchart}
    \end{figure*}
    
    %Thus, it is important to create a large database of relations, that consider \textit{relation} as a concept. Further it must cover all possible relations that could exist between a pair of entities, taxonomical and semantic association between relations, and their synsets. This study initiates work in that direction. We assume relations appearing in unstructured text resources such as Wikipedia, and structured knowledge bases (KBs) like Wikidata are a good representative of all possible relations. We extract relations from Wikipedia Infobox templates\footnote{\url{https://en.wikipedia.org/wiki/Wikipedia:List_of_infoboxes}} manually, and DBpedia and Wikidata in a data-driven way. We collect an exhaustive list of relations between \textit{person}, \textit{organization}, and \textit{location} entity types. Further, we create a relation hierarchy of 623 canonical relations. We perform analyses to understand the contribution of each resources, effects of canonicalization, complementarity of KBs and coverage of relation present in the existing RE datasets.
    %With that motivation, this work initiates study in that 
    %With that motivation, in this work, we initiate by exploiting Wikipedia Infobox templates\footnote{\url{https://en.wikipedia.org/wiki/Wikipedia:List_of_infoboxes}} manually, and DBpedia and Wikidata in a data-driven way.

   Thus, it is important to create a large database of relations, that considers \textit{relation} as a concept. Further, it must cover all possible relations that could exist between a pair of entities, taxonomical and semantic association between relations, and their synsets. This study initiates work in that direction. We assume properties and attributes appearing in structured knowledge bases (KBs) and ontologies are a good representative of all possible relations. Therefore, we extract relations from Wikipedia Infobox templates\footnote{\url{https://en.wikipedia.org/wiki/Wikipedia:List_of_infoboxes}} manually, and DBpedia and Wikidata in a data-driven way. We collect an exhaustive list of relations between \textit{person}, \textit{organization}, and \textit{location} entity types. Further, we create a relation hierarchy of 623 canonical relations. We perform analyses to understand the contribution of each resource, the effects of canonicalization, the complementarity of KBs, and coverage of relations present in the existing RE datasets.

\section{RE Datasets}
    ACE multilingual corpus \cite{mitchell2005ace} is one of the most commonly used RE dataset. It arranges relations in a hierarchy of depth 1 and contains about 30 relations at leaf level. The relations at intermediate nodes are not generic enough to be scalable and also are few in numbers.
    %The number of relations at intermediate nodes is very few and are not generic enough to be scalable. 
    \citet{mintz2009distant} proposed distant supervision for automatic data generation with more number of relations using a KB. Following that \citet{riedel2010modeling} introduced NYT dataset with 52 relations using Freebase \cite{bollacker2008freebase} as a KB. Although Freebase has more than 700 properties, only 52 could qualify as relation because of the underlying corpus. Recently published datasets TACRED \cite{zhang2017tacred}  and FewRel \cite{hao2018fewrel}, cover 42 and 100 relations respectively. Similar to our work, TACRED considers relations specific to \textit{person}, \textit{organization} and \textit{location} entity types. FewRel contains relations from Wikidata. However, the relation count is limited in contrast to our objective. 

\section{Relation Hierarchy}
    A relation triple (\texttt{e\textsubscript{1}},\texttt{r}, \texttt{e\textsubscript{2}}) represents relation \texttt{r} between head entity \texttt{e\textsubscript{1}} and tail entity \texttt{e\textsubscript{2}}. We extract triples from DBpedia and Wikidata dump. The triples are later used for finding relations and support set of relations.
    %A relation hierarchy organizes relations into groups sharing similarities. 
    %A relation \texttt{r}, between entities \texttt{e\textsubscript{1}} (head entity) and \texttt{e\textsubscript{2}} (tail entity), represents connection between them.
    
    Figure~\ref{fig:flowchart} summarizes the following steps of hierarchy creation:
    \begin{itemize}
        \item we start with extracting relations from Wikipedia Infobox templates manually, and DBpedia and Wikidata in a data-driven way between \textit{person}, \textit{organization}, and \textit{location} entity types.
        \item In step 2 we manually canonicalize relation names for smooth merging of hierarchy.
        \item In step 3, we filter relations from the list based on the frequency.
        \item In step 4, we create hierarchy for individual knowledge resource manually based upon collective judgement of all the three authors.
        \item Finally, we create a relation hierarchy of 623 canonical relations.
    \end{itemize}{}
    Above mentioned steps are discussed in details in following subsections,
    %We followed the following series of steps for creating the final hierarchy (Figure \ref{fig:flowchart}).
    \subsection{Getting relation list}
    %We chose Wikipedia Infobox, Wikidata, and DBpedia for generating a list of prospective relations.
    Wikipedia infobox stores structured information in the form of attribute-value pairs following an infobox template. Since entries in the infobox are done manually by crowd-sourced workers, we observed lots of irregularities while parsing the infobox. Therefore instead of collecting triples and relations automatically from infobox, we chose to manually scan template pages to curate a list of relations. We selected 170, 77 and 89 infobox templates for \textit{person}, for \textit{organization}, and \textit{location} respectively. We followed Wikipedia infobox template categories while selecting templates. And used translation count\footnote{Number of Wikipedia pages that uses the template} for filtering templates. We refer this list of relations as $\mathcal{L\textsubscript{i}}$.
    
    For DBpedia and Wikidata, we follow a data-driven approach. We parse Wikidata json dump\footnote{\url{https://dumps.wikimedia.org/enwiki/}} and DBpedia mapping-based Infobox dump\footnote{\url{http://downloads.dbpedia.org/2016-10/core/}} for generating triples. 
    %A triple is a tuple of \texttt{e\textsubscript{1}} (head entity), relation \texttt{r}, and \texttt{e\textsubscript{2}} (tail entity). 
    Then we collect all the unique relations from the triples dataset (where \texttt{e\textsubscript{1}} and \texttt{e\textsubscript{2}} is one of the three type \textit{person}, \textit{organization}, and \textit{location}). The two lists of relations are henceforth referred as $\mathcal{L\textsubscript{d}}$ (from DBpedia) and $\mathcal{L\textsubscript{w}}$ (from Wikidata).
        
   % We got three lists of relations \mathcal{L\textsubscript{i}}, \mathcal{L\textsubscript{d}} and \mathcal{L\textsubscript{w}} from Wikipedia Infobox templates, DBpedia and Wikidata respectively.
        
    \subsection{Name canonicalization}
    Even though the three sources are closely related, they follow different nomenclature for their relations. Thus, same relation can have different names in different lists. For example, consider relation \texttt{placeOfBirth}, it is \texttt{birth\_place} in Wikipedia Infobox , \texttt{birthPlace} in DBpedia, and \texttt{place of birth} in wikidata, .
        
    To canonicalize relation names, we follow the current policy of DBpedia. For example, if a relation name is a single word, consider it as it is, given all the characters are in small-case. Otherwise, capitalize all the words except the first word, remove in-between spaces and concatenate all the words. For example, \textit{place of birth} becomes \texttt{placeOfBirth}. In the case of multiple names for the same relation (as in the earlier example), we choose one of them and store the respective mapping. Following this procedure, we obtain canonicalized relation lists $\mathcal{C\textsubscript{i}}$, $\mathcal{C\textsubscript{d}}$ and $\mathcal{C\textsubscript{w}}$ from $\mathcal{L\textsubscript{i}}$, $\mathcal{L\textsubscript{d}}$ and $\mathcal{L\textsubscript{w}}$ respectively.
        
    \subsection{Filtering} 
    This step ensures that our relation hierarchy focuses on frequently occurring relations. We filter out relations from the list $\mathcal{C\textsubscript{i}}$ if they appear in less than 100 infoboxes. Similarly, a relation is filtered out from the lists $\mathcal{C\textsubscript{d}}$ and $\mathcal{C\textsubscript{w}}$ if it has less than 100 associated triples in their support set. 
    %Respective filtered lists are referred to as \mathcal{F\textsubscript{i}}, \mathcal{F\textsubscript{d}} and \mathcal{F\textsubscript{w}}.
    %For filtering Infobox relation, if a relation appears in less than 100 infoboxes, we discard that relation. Similarly, for filtering DBpedia and Wikidata relation, if a relation has less than 100 triples associated with it, we discard that relation.
        
   % After filtering relations from \mathcal{C\textsubscript{i}}, \mathcal{C\textsubscript{d}} and \mathcal{C\textsubscript{w}}, we get \mathcal{F\textsubscript{i}}, \mathcal{F\textsubscript{d}} and \mathcal{F\textsubscript{w}}.
       
    \subsection{Hierarchy creation}
    A relation \texttt{r} describes relationship between two entities, associating with certain entity types \citet{jain2018}. Thus, it is natural to classify based on the head and tail entity types at the top level. Consider a relation \texttt{founder}, head entity type is organization (\texttt{org}) and tail entity is of type person (\texttt{per}), thus it falls under branch \texttt{org-per}. Since it is organization specific relation, \texttt{org-per.founder} will fall under \texttt{org} which falls under the root \texttt{rel}.
        
    Initial levels of hierarchy are described as:
    \begin{itemize}
        \item At depth 0: root node referred as \texttt{rel}
        \item At depth 1: we distinguish based on head entity type. In this level there are 3 nodes \texttt{per} (person relations), \texttt{loc} (location relations), and \texttt{org} (organization relations).
        \item At depth 2: we distinguish based on both head and tail entities. In this level, there are 9 nodes (For example, \texttt{per-loc} head entity: \textit{person} and tail entity: \textit{location}) and each node of this level are henceforth referred as a bucket for relations.
    \end{itemize}
        
    All the relations from the three filtered lists 
    %\mathcal{F\textsubscript{i}}, \mathcal{F\textsubscript{d}} and \mathcal{F\textsubscript{w}}
    are distributed across 9 buckets.
    We manually arrange relations in the hierarchy whenever \textit{is-a}  association exists between two relations.
    
    %which later is manually arranged following \textit{is-a} relation between relations, if exist. 
        
    Taxonomically similar relations (child nodes) are placed under the same parent node. If the parent node is not present in the filtered relation list, the canonical relation list is referred to. If present, that referred relation is chosen. Otherwise, a new parent node is introduced. In our hierarchy, we have introduced a total of 12 new nodes.
    
    Hierarchy creation is done manually based on the collective judgment of all the authors following relation triples (collected from KBs) associated with each relation. Manual efforts ensure the hierarchy to be more interpretable and noise-free.
    
    \subsection{Hierarchy merging}Following guidelines in previous step, hierarchies $\mathcal{H\textsubscript{i}}$, $\mathcal{H\textsubscript{d}}$ and $\mathcal{H\textsubscript{w}}$ are created. Finally, they are merged into one common hierarchy $\mathcal{H}$ by eliminating the duplicates and placing taxonomically similar relation under the same branch.
        
\section{Analysis}
\begin{figure}
        \centering
        \includegraphics[width=0.5\textwidth]{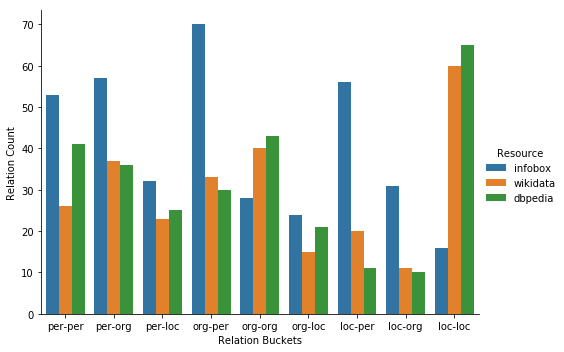}
        \caption{Distribution of relations from different resources in each of the 9 buckets.}
        \label{fig:rel_buckets}
    \end{figure}

\begin{figure}
        \centering
        \includegraphics[width=0.25\textwidth]{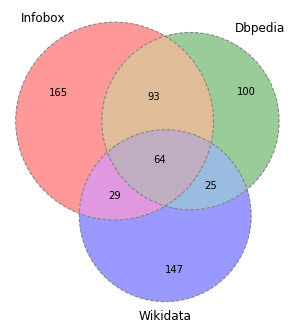}
        \caption{Contribution of resources towards common hierarchy.}
        \label{fig:rel_coverage}
    \end{figure}

    Table \ref{tab:hierarchy} shows basic statistics of three individual hierarchies $\mathcal{H\textsubscript{i}}$, $\mathcal{H\textsubscript{d}}$, $\mathcal{H\textsubscript{w}}$, and the common hierarchy $\mathcal{H}$. All hierarchies have maximum depth of 5 (6 levels).
    %We observed maximum depth 5 (6 levels) in all the hierarchies.
    All relations from the filtered lists are distributed at depths 3, 4 and 5. Distribution of relations at depth 4 and 5 gives more fine-grained information about relations shared between two entities. In the common hierarchy, \texttt{loc-loc} bucket has the most number of relations (113) whereas \texttt{org-loc} bucket has the least number (24).\\
    \begin{table}[]
        \centering
        \begin{tabular}{|c|c|c|c|c|}
            \hline
            & \begin{tabular}[c]{@{}c@{}}Relation\\ Count\end{tabular} & \begin{tabular}[c]{@{}c@{}}Relation\\ @d = 3\end{tabular} & \begin{tabular}[c]{@{}c@{}}Relation\\ @d = 4\end{tabular} & \begin{tabular}[c]{@{}c@{}}Relation\\ @d = 5\end{tabular} \\ \hline
            $\mathcal{H}$ & 623 & 357 & 247 & 19 \\ \hline
            $\mathcal{H\textsubscript{i}}$ & 351 & 177 & 168 & 6 \\ \hline
            $\mathcal{H\textsubscript{d}}$ & 282 & 162 & 110 & 10 \\ \hline
            $\mathcal{H\textsubscript{w}}$ & 267 & 209 & 52 & 6 \\ \hline
        \end{tabular}
        \caption{Relation Count of hierarchies and number of relations at various depths (Relation @ d =).}
        \label{tab:hierarchy}
    \end{table}
    \textbf{Effects of canonicalization: }Relation name canonicalization has played an important role in eliminating redundant relations from $\mathcal{L\textsubscript{i}}$ (Table \ref{tab:canonical_individual}). This, in turn, helped significantly in finding common relations among the resources. Since DBpedia and Wikidata are structured at its core, canonicalization has not affected much.\\ 
    %Still, the number of common relations between the resources has significantly increased after canonicalization.\\
    \begin{table}[]
        \centering
        \begin{tabular}{|l|l|l|l|l|l|l|}
            \hline
            & \multicolumn{2}{l|}{Person} & \multicolumn{2}{l|}{Organization} & \multicolumn{2}{l|}{Location} \\ \hline
            & B & A & B & A & B & A \\ \hline
            Infobox & 660 & 154 & 228 & 165 & 183 & 84 \\ \hline
            Dbpedia & 94 & 92 & 103 & 99 & 91 & 86 \\ \hline
            Wikidata & 71 & 71 & 73 & 72 & 98 & 97 \\ \hline
        \end{tabular}
        \caption{Relation counts before (B) and after (A) canonicalization.}
        \label{tab:canonical_individual}
    \end{table}
    \textbf{Coomplementarity of resources: }Figure \ref{fig:rel_buckets} shows the contribution of resources towards relation buckets. Manually collected relations from Wikipedia Infobox dominate 7 out of the 9 buckets. The contributions of DBpedia and Wikidata towards each bucket is almost similar.\\
    Figure \ref{fig:rel_coverage} shows the contribution of each resource towards the common hierarchy. Only about $10\%$ relations are common among the three resources. This analysis indicates the complementarity of the resources.\\
    %Manually extracted relations contributes the most, 165 relations. Together all three resources contribute 64 relations. When we consider two resources, DBpedia and manually extracted relations combined contribute the, 93 relations. (Figure \ref{fig:rel_coverage})\\
    \textbf{Comparison with relation list of RE datasets: }The main objective behind this study was to highlight the major bottlenecks of RE datasets (sec~\ref{sec:intro}). Table \ref{tab:re-datasets} briefly shows how relations from RE datasets get subsumed in our relation hierarchy. Our hierarchy covers an average of 62\% of relations when all the relation of a dataset is considered and 85.35\% of relations when relation's head and tail entity types are restricted to \textit{person}, \textit{organization}, and \textit{location} types.\\
    \begin{table}[]
        \centering
        \begin{tabular}{|p{1.8cm}|p{1.2cm}|p{1.2cm}|p{1.8cm}|}
            \hline
            RE Dataset & relation count & depth & relation subsumed  \\ \hline
            ACE 2004 & 24 (17) & 1 & 11   \\ \hline
            NYT2010 Dataset & 51 (47) & 0 & 35  \\ \hline
            TACRED & 41 (29) & 0 & 27  \\ \hline
            FewRel & 100 (64) & 0 & 61  \\ \hline
            Relation Hierarchy & 623 & 5 (3.42) & - \\ \hline
        \end{tabular}
        \caption{RE Datasets with relation count(relation with head and tail entity of type \textit{person}, \textit{organization}, and \textit{location}), hierarchy depth, and numbers of relation subsumed in Relation Hierarchy}
        \label{tab:re-datasets}
    \end{table}
    
\section{Use Cases and Applications}
    \textbf{RE dataset creation: }We provide a comprehensive list of 623 relations. This list, along with relation hierarchy, can further be used as a set of model relations for creating a sizeable sentence-level dataset for RE.\\
    \textbf{Improving RE training data: }Introduction of relation hierarchy will also guarantee training data for intermediate nodes. This will help solve the problem for some of the long-tail labels. For example, in TACRED, percentage of training instances for \texttt{city\_of\_death},  \texttt{country\_of\_death}, and  \texttt{stateorprovince\_of\_death} are 0.12\%, 0.01\% and 0.07\% respectively which is significantly less than average, 0.46 \% (excluding no\_relation). In the relation hierarchy, these relations will be placed under \texttt{place\_of\_death}. This new relation will contain instances of all the three (combined percentage 0.20\%).\\
    \textbf{Hierarchical RE Models: }\citet{han2018hierarchical} proposed a hierarchical RE model on NYT2010 dataset. Their hierarchical model significantly outperformed other baselines by utilizing Freebase hierarchy. Our relation hierarchy being more comprehensive and scalable, we expect better learning for hierarchical RE models.\\
    \textbf{Knowledge Graph Completion: }Despite the large size of knowledge bases, they are far from complete. For example, only 25\% of Dbpedia \textit{person} and 12\% of Wikidata \textit{person} has \texttt{placeOfBirth} information, and 27\% of Dbpedia \textit{location} and 24\% of Wikidata \textit{location} has \texttt{country} information. Once we map relations from multiple resources to the canonicalized relations of relation hierarchy, we can easily compare the triples of different knowledge resources.\\
    
\section{Conclusion and Future Work}
    This study explored more than 1500 properties and attributes from Wikipedia Infoboxes, DBpedia, and Wikidata to generate lists of prospective relations. These relations were used to create a hierarchy of 623 canonical relations. Our analysis indicates only a 10\% overlap among the three resources. Additionally, our relation hierarchy subsumes 85\% of relations from RE datasets with restricted entity types. In future work, we aim to extend this relation hierarchy by including more entity types, and more resources like YAGO \cite{suchanek2007yago}. We want to extend the hierarchy in an automated manner to increase the coverage of resources. Furthermore, we also intend to use this extensive list of relations along with the relation hierarchy for generating a large-scale dataset for fine-grained RE.

\bibliography{emnlp-ijcnlp-2019}
\bibliographystyle{acl_natbib}

\end{document}